%% file: main.tex
\documentclass{article}
\usepackage{spconf,amsmath,graphicx}
\usepackage{booktabs}
\usepackage{subcaption}
\usepackage{amsfonts}
\usepackage{bm}
\usepackage{url}
\usepackage{hyperref}
\usepackage{xcolor}

\title{Generalized nested latent variable models for lossy coding applied to wind turbine scenarios}

\name{Raül~Pérez-Gonzalo$^{1,2}$, Andreas~Espersen$^{2}$ and Antonio~Agudo$^{1}$\thanks{This work has been supported by the Innovation Fund Denmark under 2021 ID1044-0044A and by the project MoHuCo PID2020-120049RB-I00 funded by MCIN/AEI/10.13039/501100011033.}}
\address{$^{1}$Institut de Robòtica i Informàtica Industrial, CSIC-UPC, Spain\\$^{2}$Wind Power LAB, Copenhagen, Denmark}
\begin{document}
%
\maketitle
\begin{abstract}
Rate-distortion optimization through neural networks has accomplished competitive results in compression efficiency and image quality. This learning-based approach seeks to minimize the compromise between compression rate and reconstructed image quality by automatically extracting and retaining crucial information, while discarding less critical details. A successful technique consists in introducing a deep hyperprior that operates within a $2$-level nested latent variable model, enhancing compression by capturing complex data dependencies. This paper extends this concept by designing a generalized $L$-level nested generative model with a Markov chain structure. We demonstrate as $L$ increases that a trainable prior is detrimental and explore a common dimensionality along the distinct latent variables to boost compression performance. As this structured framework can represent autoregressive coders, we outperform the hyperprior model and achieve state-of-the-art performance while reducing substantially the computational cost. Our experimental evaluation is performed on wind turbine scenarios to study its application on visual inspections. 
\end{abstract}

\input{Chapters/Introduction}

\input{Chapters/Methods}
\input{Chapters/Results}
\input{Chapters/Conclusion}

\vspace{-0.35cm}


\bibliographystyle{IEEEbib}
\bibliography{References}

\end{document}

%% file: Chapters/Introduction.tex
\vspace{-0.25cm}
\section{Introduction}
\vspace{-0.25cm}

Lossy image compression aims to convert an image to a compressed representation by capturing its spatial redundancies. It sacrifices the ability to perfectly reconstruct the original data to achieve a higher compression ratio. In response to the ever-growing demand for efficient image storage and transmission, distinct traditional codecs have emerged. Notably, these routines are grounded in manually tailored algorithms: JPEG2000~\cite{J2K} uses a wavelet-based transform, WebP~\cite{webp} combines predictive coding and discrete cosine transform, HEVC~\cite{hevc} integrates spatial prediction and transform coding, BPG~\cite{bpg} employs context modeling, and VTM~\cite{vtm} utilizes intra and inter prediction, and block-based transform coding.

Instead of relying on handcrafted algorithms, learning-based codecs formulate image compression as an optimization problem~\cite{review,review2}. By implementing a relaxed rate-distortion loss~\cite{rnn,review3}, they can directly generate compressed representations of data and automatically learn to prioritize what information to retain for optimal compression performance. 

\begin{figure}[t!]
  \centering
  \includegraphics[width=8.6cm]{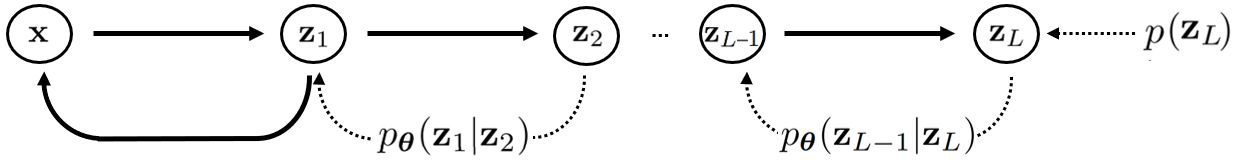}
  \vspace{-0.5cm}
\caption{\textbf{Generalized nested latent variable model for lossy compression.} Solid arrows signify direct calculations of the latent variables $\mathbf{z}_l$ from the encoder and the input image $\mathbf{x}$ from the decoder, while dashed arrows entail the estimation of likelihood and prior distributions.}
\label{fig:nested_model}
\vspace{-0.65cm}
\end{figure}

A common powerful technique involves leveraging a latent variable, a concealed factor not inherently part of the input data, which strengthens the model's capacity to capture intricate data dependencies. Bls2017~\cite{bls2017} proposes an end-to-end optimized image compression framework that utilizes an autoencoder network and a factorized trainable prior. HP~\cite{hyperprior} extends this set-up by introducing a hyperprior, adding a second latent variable nested to the previous one. Recent advances in this field explore a coarse-to-fine architecture for enhancing conditional entropy modeling~\cite{l3}. An extended coarse-to-fine approach with ResNet architecture is presented in~\cite{wacv}, and QARV~\cite{qarv} further refines it with variable-rate compression by embedding the specific rate-distortion trade-off desired through adaptive normalization. Other approaches apply a context prediction module as JA~\cite{ja} or others~\cite{context,context2,context_icassp,mlic}, tailored non-linear transforms~\cite{easn} or attention mechanisms~\cite{attention} such as GMM-Anchor/-Attn~\cite{gmm,jiro}.


In this work, we extend the hyperprior concept by stacking $L$ distinct layers of latent variable~\cite{deep-latent1,deep-latent2}, as illustrated in Fig.~\ref{fig:nested_model}. Therefore, we construct a generalized nested latent model with a more flexible conditional entropy model, which is easily parallelizable~\cite{time}. We demonstrate that the optimal level of nested layers depend on the rate-distortion trade-off desired. In addition to that, we showcase that for greater levels of $L$ a trainable prior is detrimental, thus, we explore a predefined logistic prior along with a common dimension for the latent variables to enhance generalized nested models.

Our approach is evaluated on a real-world industry problem. In particular, we study generalized nested models on wind turbine imagery captured during blade inspections~\cite{PerezGonzaloIcip2023,ai-drone}. Reducing the image size without compromising its quality is crucial to properly assess wind turbines and arrange their repair~\cite{downtime,downtime2}. We showcase nested latent variables models can successfully approximate autoregressive models and, therefore, generalize them. Hence, our approach reaches competing performance compared to state-of-the-art coders on the presented challenging data, while substantially reducing its running cost, thus, it emerges as the optimal lossy coding solution for wind industry applications.




%% file: Chapters/Methods.tex
\vspace{-0.25cm}
\section{Nested Latent Variable Models for Lossy Coding}
\label{section_methods}
\vspace{-0.2cm}

We aim to learn a probabilistic model $p_{\boldsymbol{\theta}}(\mathbf{x})$ of our observed data $\mathbf{x}$ to successfully apply an entropy coder capable of compressing them, where $ \boldsymbol{\theta}$ denotes the model parameters. To address this learning problem, fully-observed models can be marginalized over a latent variable $\mathbf{z}_1$. Let us denote the likelihood distribution $p_{\boldsymbol{\theta}} (\mathbf{x} | \mathbf{z}_1)$ and the prior distribution $p(\mathbf{z}_1)$, then, by the Bayes' rule, $p_{\boldsymbol{\theta}}(\mathbf{x})$ can be expressed as the joint distribution of the observed and latent variables as:

\vspace{-0.4cm}
\begin{equation} \label{eq:bayes}
    p_{\boldsymbol{\theta}} (\mathbf{x}) = \int p_{\boldsymbol{\theta}} (\mathbf{x}, \mathbf{z}_1) d\mathbf{z}_1 
  = \int p_{\boldsymbol{\theta}} (\mathbf{x} | \mathbf{z}_1) p(\mathbf{z}_1) d\mathbf{z}_1 .
\end{equation}
\vspace{-0.4cm}

This implicit representation can be recursively applied to each latent variable $\mathbf{z}_l$ to obtain an $L$-layer Markov model (see Fig.~\ref{fig:nested_model}), where $l \in \{1, \ldots, L\}$. In this way, we can sequentially gather the distinct latent dependencies to express the model evidence $ p_{\boldsymbol{\theta}} (\mathbf{x})$ defined in Eq.~\eqref{eq:bayes} as:

\vspace{-0.35cm}
\begin{equation} \label{eq:conditional}
    p_{\boldsymbol{\theta}} (\mathbf{x}) = \int p_{\boldsymbol{\theta}} (\mathbf{x} | \mathbf{z}_1) p_{\boldsymbol{\theta}} (\mathbf{z}_{1:L-1} | \mathbf{z}_{L}) p (\mathbf{z}_L)  
    \ d\mathbf{z}_{1:L} ,
\end{equation}
where we have defined for conciseness $p_{\boldsymbol{\theta}} (\mathbf{z}_{1:L-1} | \mathbf{z}_{L}) = \prod_{l=1}^{L-1} p_{\boldsymbol{\theta}} (\mathbf{z}_{l} | \mathbf{z}_{l+1})$ and $d\mathbf{z}_{1:L} =  d\mathbf{z}_1 d\mathbf{z}_2 \ldots  d\mathbf{z}_{L-1} d\mathbf{z}_L $.

\begin{figure}[t!]
\centering \includegraphics[width=3.3in]{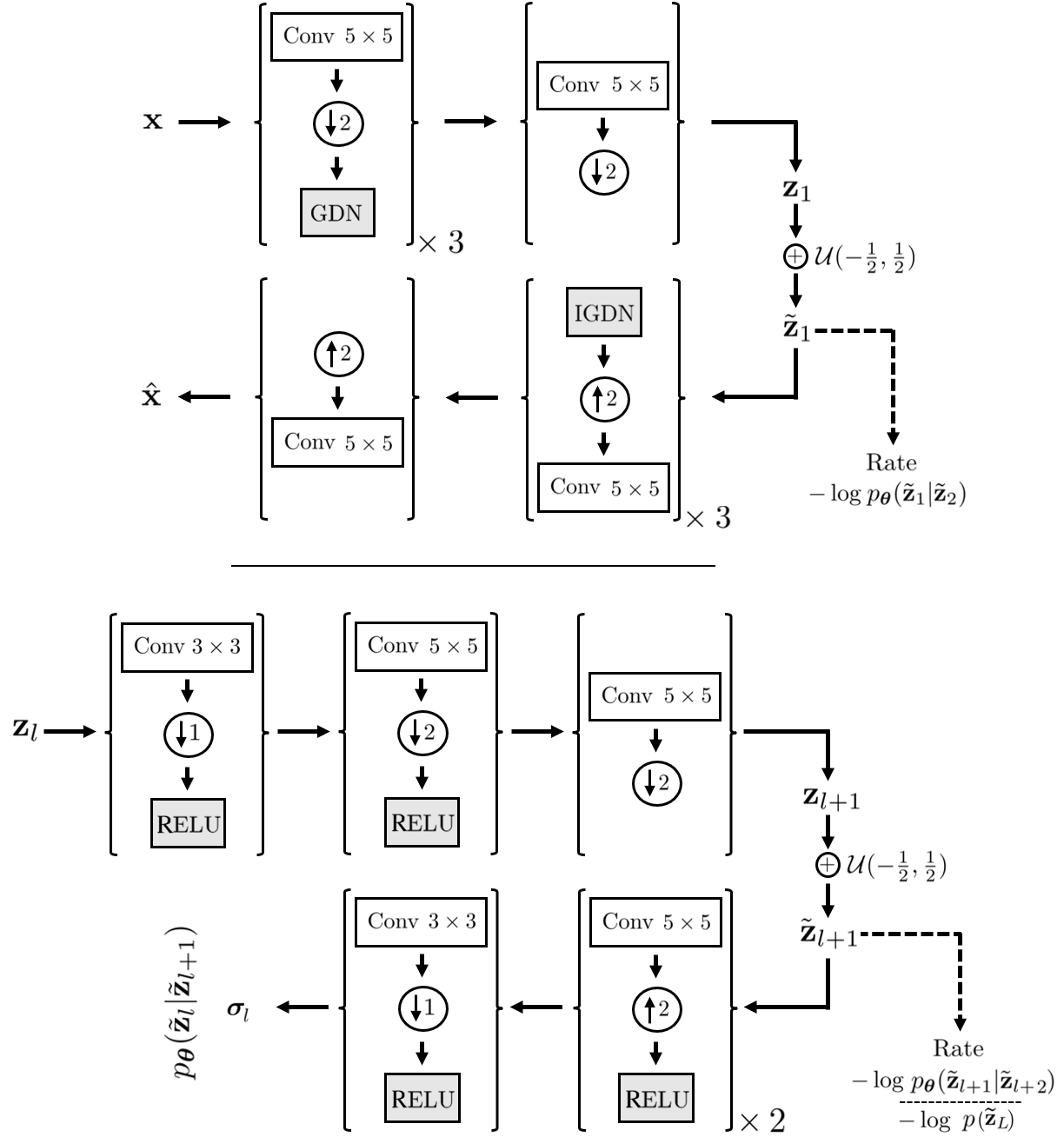}
\caption{ \textbf{Proposed architecture for a generalized nested latent variable model}. In the first layer, the decoder reconstructs directly the input image $\mathbf{x}$, while it estimates the likelihood distributions in the rest of the layers. The network is built with building blocks that are composed of a convolution, a down/upsampling operation and a nonlinear function.} \label{fig:architecture} \vspace{-0.3cm}
\end{figure}

\vspace{-0.15cm}
\subsection{Generalized Relaxed Rate-distortion Loss} \label{sec:loss}

We seek to minimize the trade-off governed by $\lambda \in \mathbb{R}$ between the compression rate and the decompression error:

\vspace{-0.4cm}
\begin{equation*} 
 - \mathbb{E}_{p_{\boldsymbol{\theta}}(\mathbf{x})} \left[ \log p_{\boldsymbol{\theta}} (\mathbf{z}_{1:L-1} | \mathbf{z}_{L}) + \log p(\mathbf{z}_{L}) \right] + \lambda \ \mathbb{E}_{p_{\boldsymbol{\theta}}(\mathbf{x})} \left[  d(\mathbf{x}, \hat{\mathbf{x}}) \right]   ,
\end{equation*}
where $d(\mathbf{x}, \hat{\mathbf{x}})$ indicates a distortion metric between the input image $\mathbf{x}$ and the reconstructed one $\hat{\mathbf{x}}$.

The presented loss operates in the continuous space, however, entropy coders can only operate with finite discrete alphabets, requiring a quantization step. In our case, we define the quantizer $Q$ as a rounding operation that is continuously approximated through uniform random noise~\cite{bls2017}. Let $\tilde{\mathbf{z}}_l = Q(\mathbf{z}_l)$ denote the noisy approximation, then we end up with the following trainable rate-distortion loss:

\vspace{-0.4cm}
\begin{equation*}  \label{eq:relaxed-rate-distortion}
 - \mathbb{E}_{p_{\boldsymbol{\theta}}(\mathbf{x})} \left[ \log p_{\boldsymbol{\theta}} (\tilde{\mathbf{z}}_{1:L-1} | \tilde{\mathbf{z}}_{L}) + \log p(\tilde{\mathbf{z}}_{L}) \right] + \lambda \ \mathbb{E}_{p_{\boldsymbol{\theta}}(\mathbf{x})} \left[  d(\mathbf{x}, \hat{\mathbf{x}}) \right]   .
\end{equation*}
\vspace{-0.4cm}

Note that the quantizer also compromises the distortion term, because $\hat{\mathbf{x}}$ relies on the latent variable $\tilde{\mathbf{z}}_{1}$. 

\vspace{-0.25cm}
\subsection{Model Architecture} \label{sec:model}
\vspace{-0.1cm}

We design our encoder and decoder with symmetrical structures, ensuring that the output dimension matches the input dimension. Both transforms employ a sequence of three blocks for each latent layer: a convolutional layer, a down/upsampling operation, and a non-linear activation function. For the initial layer, we utilize the generalized divisive normalization with adaptable parameters as the chosen non-linear function~\cite{bls2017}. Subsequent layers employ ReLU. 

Specifically, convolutions employ a 2-dimensional kernel size of $5\times5$, except those on top of the latent layers which operate with a $3$-kernel. The padding and stride are meticulously selected to ensure congruence between input and output dimensions. To minimize computational time, down/upsampling operations are seamlessly integrated with linear convolutions through adjusting the convolution stride. These convolutions utilize $70$ filters, except for those interacting with $\mathbf{z}_l$, which employ 150. A comprehensive illustration of the network architecture is presented in Fig.~\ref{fig:architecture}.

The standard logistic distribution with statistical independence across its components serves as the prior $p(\tilde{\mathbf{z}}_L)$. The likelihood $p_{\boldsymbol{\theta}} (\tilde{\mathbf{z}}_{l} | \tilde{\mathbf{z}}_{l+1})$ for $l\in \{1, \ldots, L-1\}$ are characterized as conditional independent zero-mean Gaussian distributions. Hence, these distributions are modelled by a convolutional network for every standard deviation $\{\boldsymbol{\sigma}_l\}_{l \in \{1,\ldots,L-1\}} $. 

\newpage
\vspace{-0.15cm}
\subsection{Compression Scheme} \label{sec:compression-scheme}

Each latent variable $\mathbf{z}_l = (z_{l,1}, \ldots, z_{l,2^P})$ undergoes independent discretization into $2^P$ distinct bins, where $P=10$ is a predefined precision. Within each latent layer and for every component $z_{l,j}$, the bins exhibit uniform and identical widths. Defining $B(z_{l,j})$ as the bin encompassing $z_{l,j}$, we establish the discretized value for $z_{l,j}$ as the midpoint of bin $B(z_{l,j})$. Consequently, the discretized distributions become highly smooth, particularly as the number of discretization bins increases. These distributions are combined with asymmetric numeral systems for entropy coding~\cite{ans}.

\vspace{-0.15cm}
\subsection{Autoregressive Universal Approximators} \label{sec:approximate}



Let $x^{(t)}$ denote the $t$-th pixel component of $\mathbf{x}$, autoregressive models~\cite{pixelcnn} leverage the previously decoded components to enhance the coding performance reconstruction of the next pixel component in order to learn $p_{\boldsymbol{\theta}}(\mathbf{x})$ from Sec.~\ref{section_methods}. 

\vspace{-0.3cm}
\begin{equation} \label{eq:autoregressive} p_{\theta }(\mathbf{x}) = \prod p_{\theta }(x^{(t)} | x^{(t-1)},\ldots,x^{(1)}),  \end{equation}
where $x^{(t-1)}, x^{(t-2)}\ldots, x^{(1)}$ are the previously decoded pixel components from $\mathbf{x}$.

Without loss of generality, we assume each latent variable $z_l$ only has a single component. Nested latent variable models can approximate autoregressive models by simply decoding each pixel component through an additional latent variable. In this way, we can reproduce the sequential decoding order of an autoregressive model: the first pixel component $x_1$ is decoded by $z_L$ through $p(z_L)$, the second component $x_2$ is decoded by $z_{L-1}$ through $p(z_{L-1}|z_{L})$, and so on. In general terms, the $t$-th pixel component $x^{(t)}$ is decoded by $z_{L-t+1}$ through $p(z_{L-t+1}|z_{L-t+2},\ldots,z_{L})$. Hence, the model evidence $p_{\boldsymbol{\theta}}(\mathbf{x})$ in Eq.~\eqref{eq:autoregressive} can be expressed as: 

\begin{equation} p_{\theta }(\mathbf{x}) = \prod p(z_{L-t+1}|z_{L-t+2},\ldots,z_{L}).  \end{equation}
The provided rationale can be applied to approximate any autoregressive variation, including channel-wise~\cite{channel} or checkerboard~\cite{checkerboard} autoregressive models. A thorough proof is detailed in ~\cite{approximate}.

Autoregressive models, while formally capable of conditioning predictions on all previous decoded components, commonly utilize a fixed context, such as $5\times5$ convolution kernels~\cite{ja}. Thus, autoregressive lossy models can be approximated by incorporating a limited number of 
$z_{l}$ variables.

\begin{figure}[!t]
  \centering
\resizebox{8.4 cm}{!} {
  \centerline{\includegraphics[width=8.5cm]{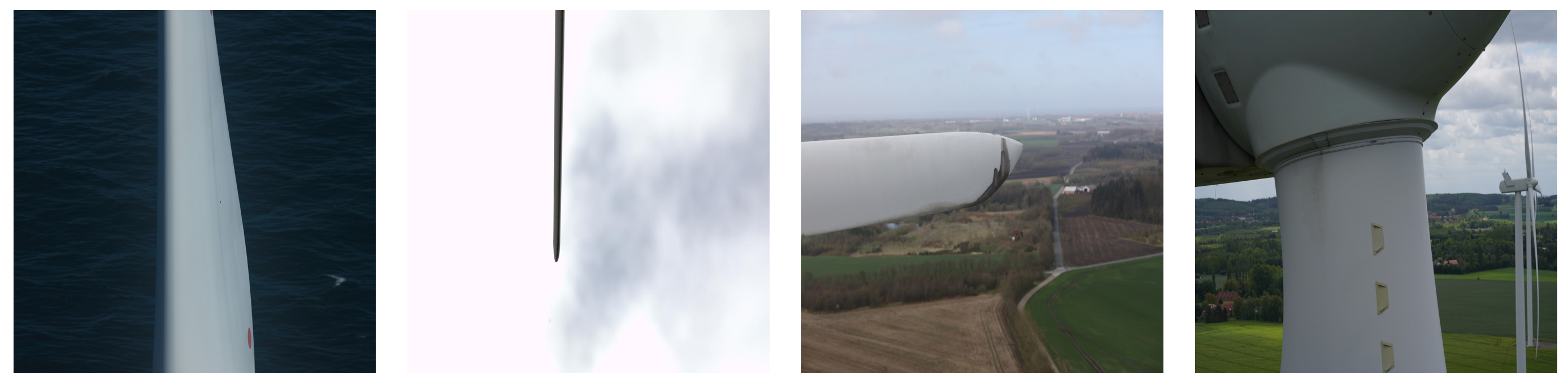}}}
  \vspace{-0.3cm}
\caption{\textbf{Four instances of wind turbine blade images.} The pictures showcase distinct blade surfaces and locations with respect to the rotor of the turbine.}
\label{fig:data}
\vspace{-0.5cm}
\end{figure}

%% file: Chapters/Results.tex
\section{Experimental Results} \label{sec:results}
\vspace{-0.3cm}

\subsection{Dataset} \label{sec:data} 
\vspace{-0.15cm}

We possess a collection of 64,438 high-resolution blade images in raw format, each with an RGB resolution of 6,744$\times$4,502 pixels (see examples in Fig.~\ref{fig:data}). These images are categorized into three sets: a training set encompassing 
$\sim$80\% of the data, along with validation and test sets, each containing $\sim$10\%. To ensure unbiased performance and effective generalization for new acquired data, images from the same inspection campaign are exclusively present in one set. During training, random crop selections are utilized. 

\begin{figure}[t!]
  \begin{minipage}{0.5\textwidth} 
    \centering
    \begin{minipage}{0.5\textwidth}
      \centering
      \includegraphics[width=1.0\linewidth]{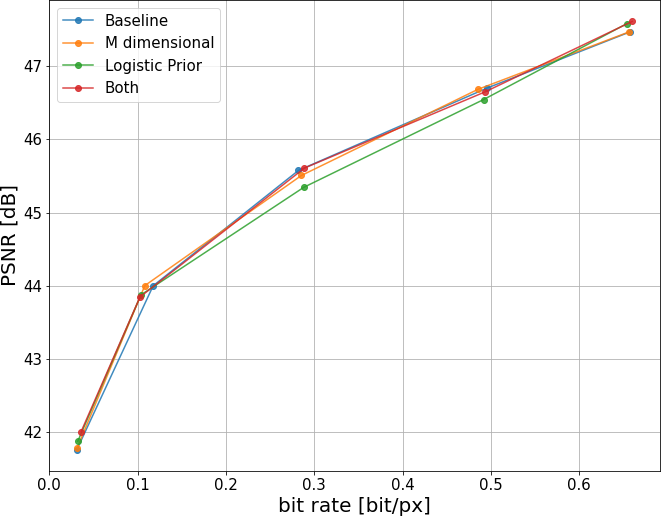}
      \vspace{-0.5cm}
      \subcaption{$L=3$}
      \label{subfig:ablation-hp3}
    \end{minipage}%
    \begin{minipage}{0.5\textwidth}
      \centering
      \includegraphics[width=1.0\linewidth]{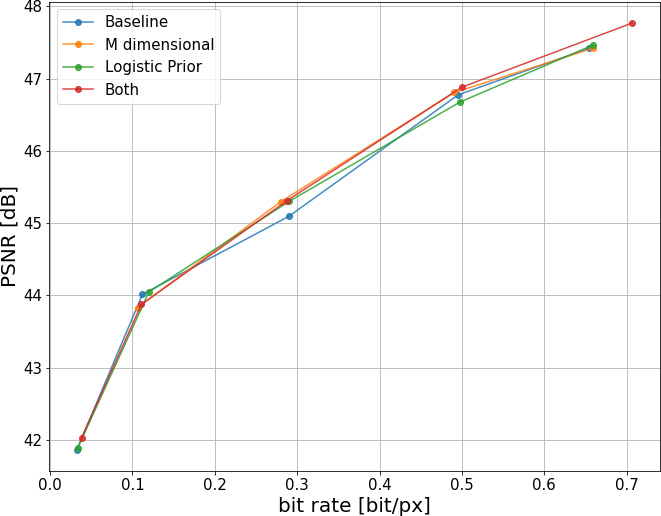}
      \vspace{-0.5cm}
      \subcaption{$L=4$}
      \label{subfig:ablation-hp4}
    \end{minipage}
    \vspace{\baselineskip} 
    \begin{subfigure}{\textwidth}
      \centering
      \includegraphics[width=0.51\linewidth]{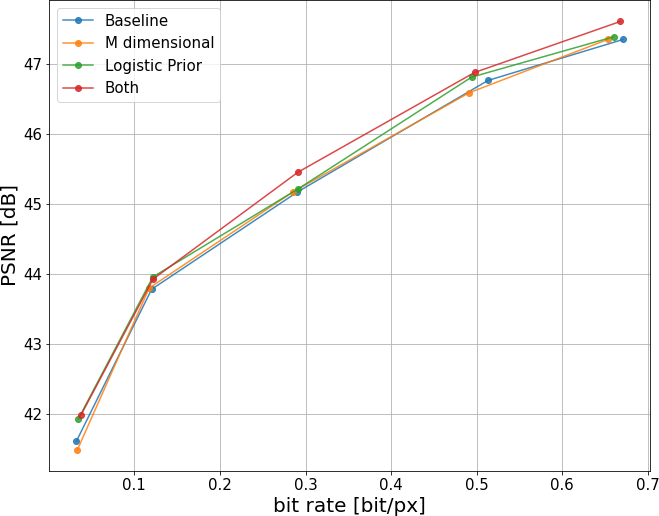} 
      \vspace{-0.1cm}
      \subcaption{$L=5$}
      \label{subfig:ablation-hp5}
    \end{subfigure}
  \end{minipage}
  \vspace{-0.75cm}
  \caption{\textbf{Ablation study results}. ``Baseline" extends exactly the architecture from~\cite{hyperprior}; ``M dimensional" denotes establishing all latents $\mathbf{z}_l$ as M-dimensional variables; ``Logistic prior" indicates employing a standard logistic for $p(\mathbf{z}_L)$.}
  \label{fig:ablation}
  \vspace{-0.3cm}
\end{figure}

\begin{figure}[t!]
  \begin{minipage}{0.5\textwidth} 
    \centering
    \begin{minipage}{0.5\textwidth}
      \centering
      \includegraphics[width=1.0\linewidth]{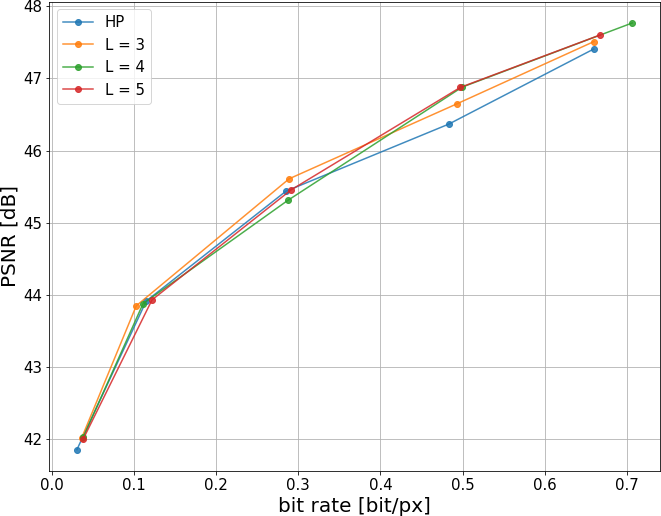}
      \vspace{-0.5cm}
      \subcaption{PSNR quality measure}
      \label{subfig:model_l-psnr}
    \end{minipage}%
    \begin{minipage}{0.5\textwidth}
      \centering
      \includegraphics[width=1.0\linewidth]{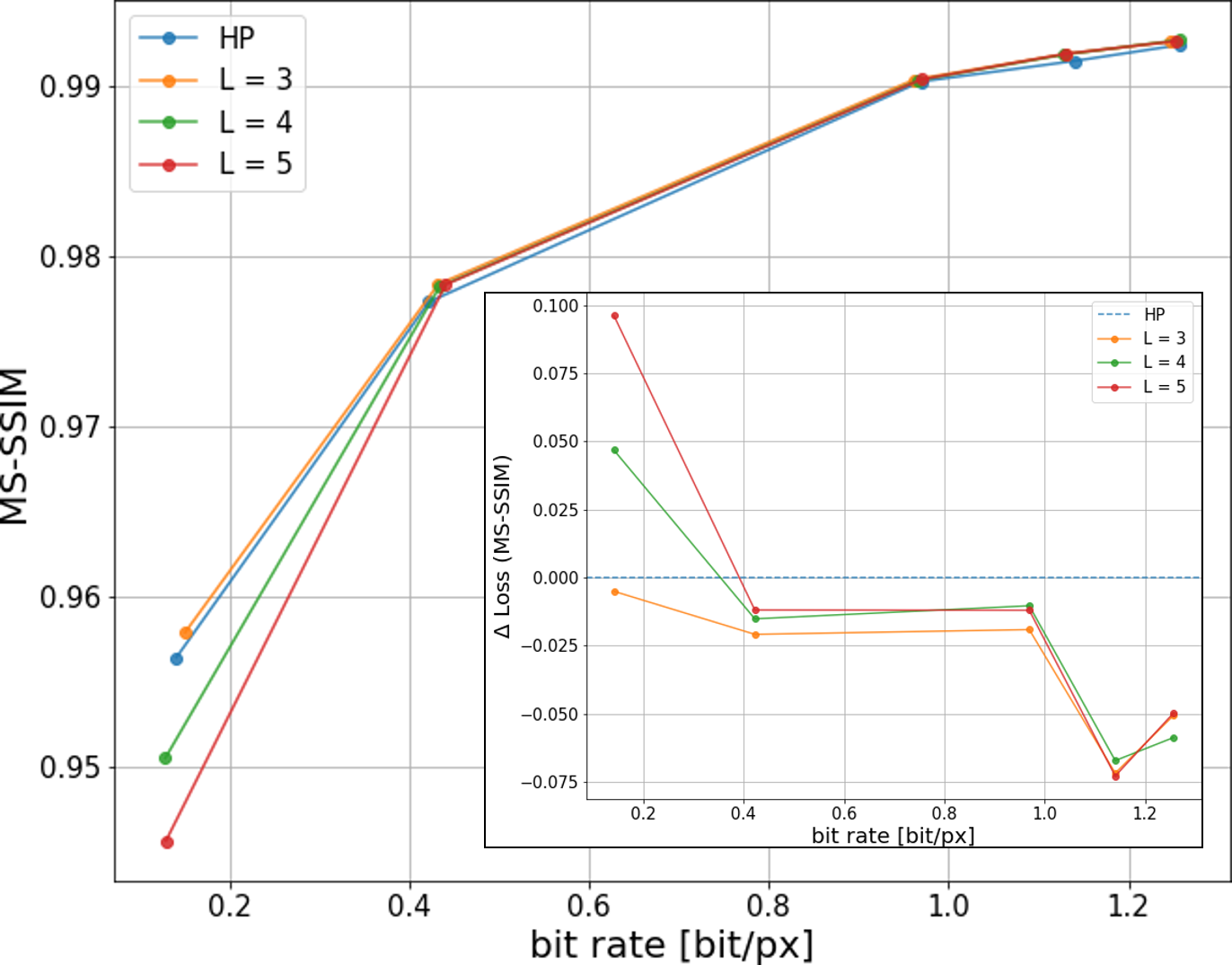}
      \vspace{-0.5cm}
      \subcaption{MS-SSIM quality measure}
      \label{subfig:model_l-msssim}
    \end{minipage}
  \end{minipage}
  \vspace{-0.3cm}
  \caption{\textbf{Quantitative performance comparison with respect to $L$ over the validation set}. HP~\cite{hyperprior} employs $L=2$. Plots are divided based on the training's distortion metric. MS-SSIM graphic contains another plot comparing loss differences between each $L$ to HP~\cite{hyperprior}.}
  \label{fig:model_l}
  \vspace{-0.5cm}
\end{figure}


\begin{figure*}[ht!]
\begin{center}
\resizebox{17.8cm}{!} {
\begin{tabular}{@{}c@{}c@{}c@{}c@{}c@{}}
\begin{subfigure}[b]{0.75\textwidth}
\includegraphics[width=\linewidth]{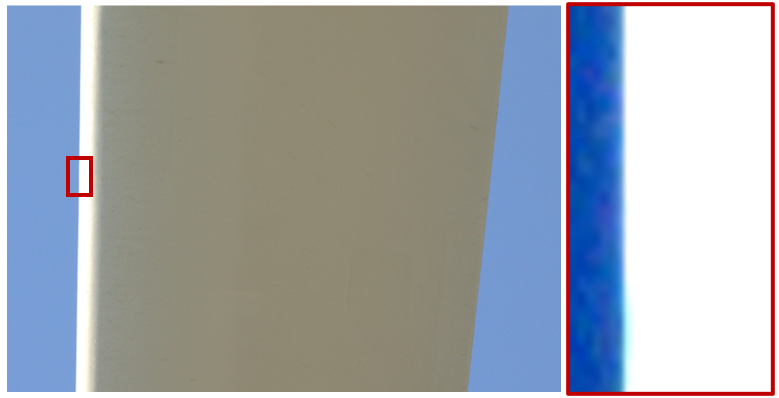}
\caption*{\centering \parbox{\linewidth}{\centering\fontsize{30}{36}\selectfont Original image \\ \textcolor{white}{.} }}
\end{subfigure}
&
\begin{subfigure}[b]{0.75\textwidth}
\includegraphics[width=\linewidth]{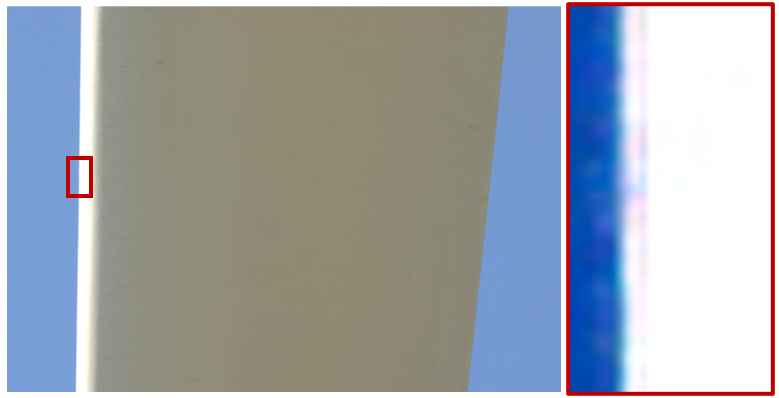}
\caption*{\centering \parbox{\linewidth}{\centering\fontsize{30}{36}\selectfont HP~\cite{hyperprior} \\ 40.47 / 0.9873 / 1.1238 }}
\end{subfigure}
&
\begin{subfigure}[b]{0.75\textwidth}
\includegraphics[width=\linewidth]{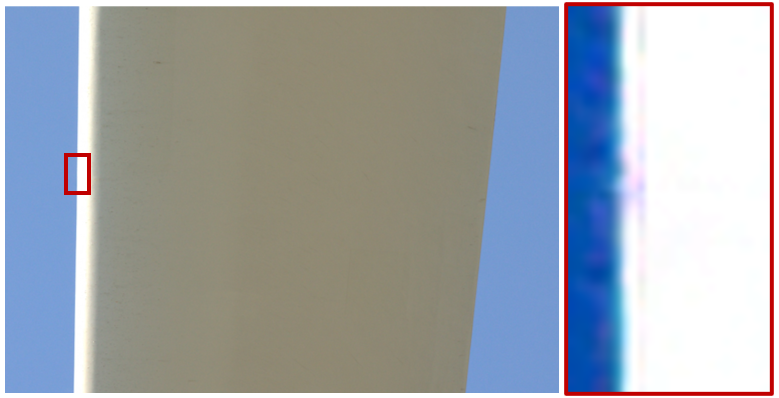}
\caption*{\centering \parbox{\linewidth}{\centering\fontsize{30}{36}\selectfont $L=3$ \\ 40.52 / 0.9874 / 1.1294 }}
\end{subfigure}
&
\begin{subfigure}[b]{0.75\textwidth}
\includegraphics[width=\linewidth]{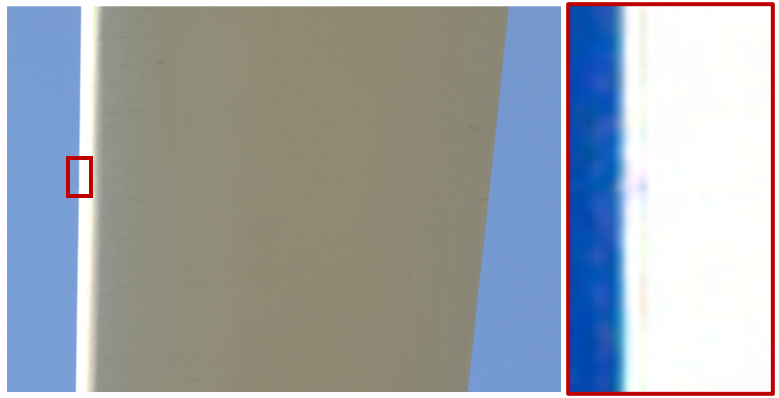}
\caption*{\centering \parbox{\linewidth}{\centering\fontsize{30}{36}\selectfont $L=4$ \\ 40.67 / 0.9878 / 1.1405 }}
\end{subfigure}
&
\begin{subfigure}[b]{0.75\textwidth}
\includegraphics[width=\linewidth]{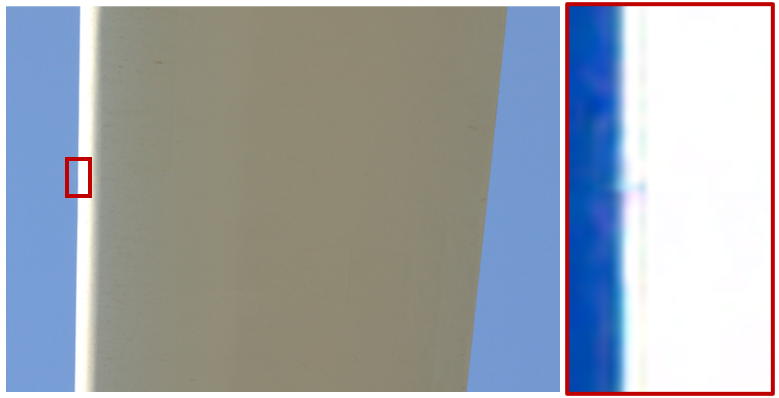}
\caption*{\centering \parbox{\linewidth}{\centering\fontsize{30}{36}\selectfont $L=5$ \\ 40.66 / 0.9878 / 1.1403 }}
\end{subfigure}

\\ & & & & \\

\begin{subfigure}[b]{0.75\textwidth}
\includegraphics[width=\linewidth]{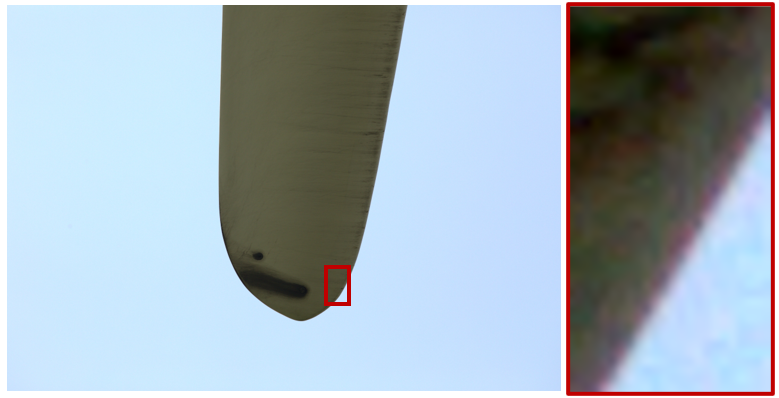}
\caption*{\centering \parbox{\linewidth}{\centering\fontsize{30}{36}\selectfont Original image \\ \textcolor{white}{.} }}
\end{subfigure}
&
\begin{subfigure}[b]{0.75\textwidth}
\includegraphics[width=\linewidth]{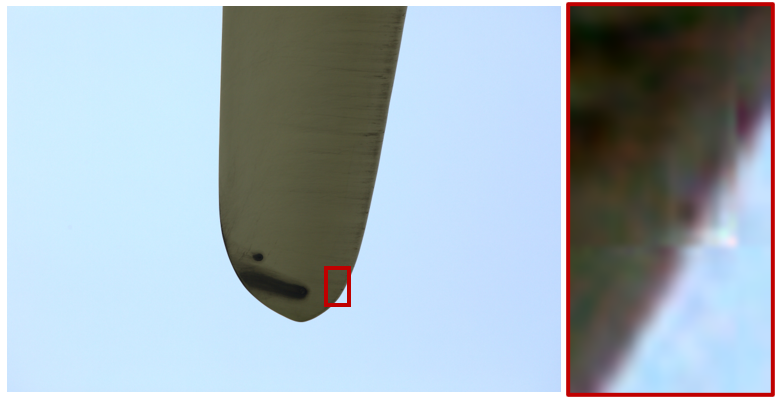}
\caption*{\centering \parbox{\linewidth}{\centering\fontsize{30}{36}\selectfont HP~\cite{hyperprior} \\ 41.55 / 0.9899 / 1.0645 }}
\end{subfigure}
&
\begin{subfigure}[b]{0.75\textwidth}
\includegraphics[width=\linewidth]{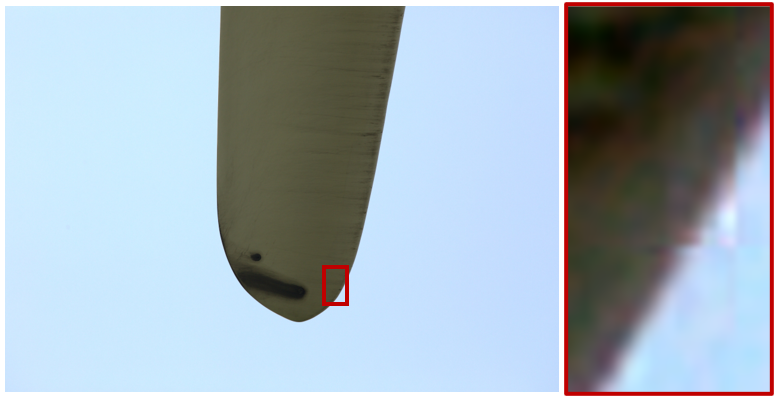}
\caption*{\centering \parbox{\linewidth}{\centering\fontsize{30}{36}\selectfont $L=3$ \\  41.60 / 0.9900 / 1.0742 }}
\end{subfigure}
&
\begin{subfigure}[b]{0.75\textwidth}
\includegraphics[width=\linewidth]{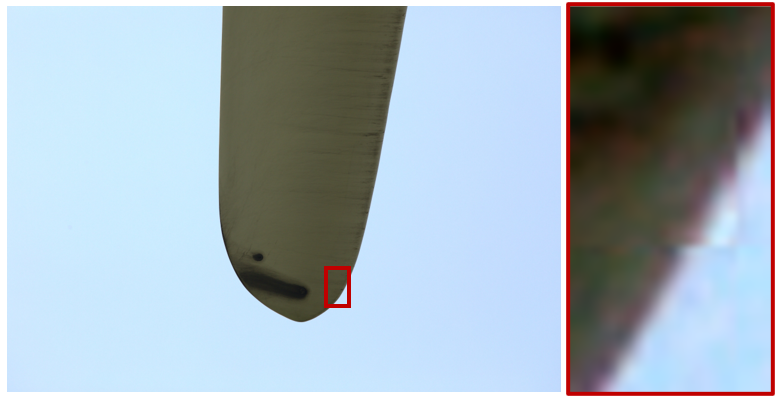}
\caption*{\centering \parbox{\linewidth}{\centering\fontsize{30}{36}\selectfont $L=4$ \\ 41.64 / 0.9900 / 1.0670 }}
\end{subfigure}
&
\begin{subfigure}[b]{0.75\textwidth}
\includegraphics[width=\linewidth]{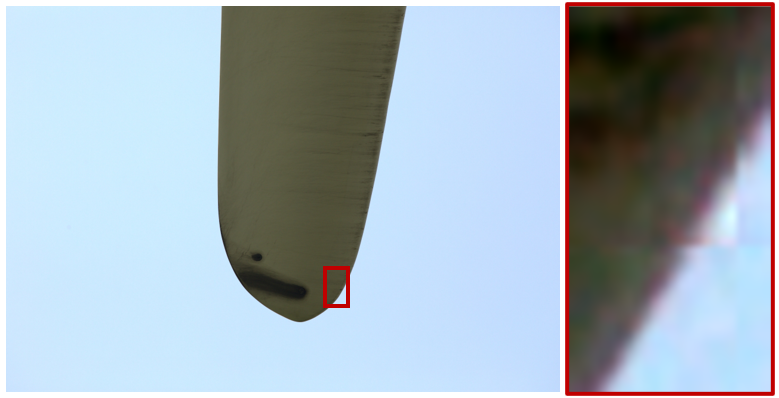}
\caption*{\centering \parbox{\linewidth}{\centering\fontsize{30}{36}\selectfont $L=5$ \\ 41.63 / 0.9900 / 1.0674 }}
\end{subfigure}

\\ & & & & \\

\begin{subfigure}[b]{0.75\textwidth}
\includegraphics[width=\linewidth]{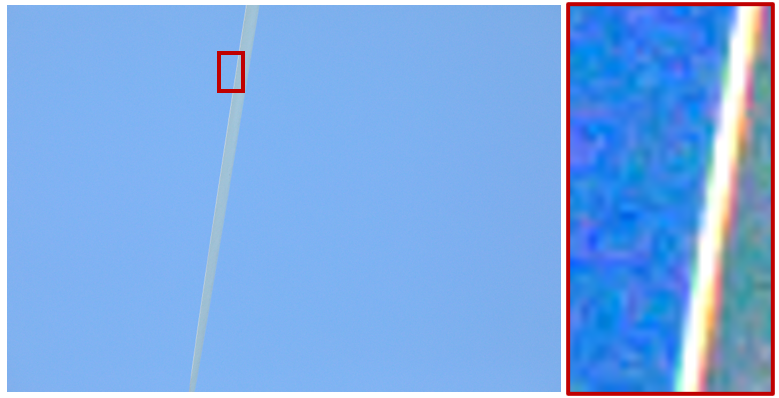}
\caption*{\centering \parbox{\linewidth}{\centering\fontsize{30}{36}\selectfont Original image \\ \textcolor{white}{.} }}
\end{subfigure}
&
\begin{subfigure}[b]{0.75\textwidth}
\includegraphics[width=\linewidth]{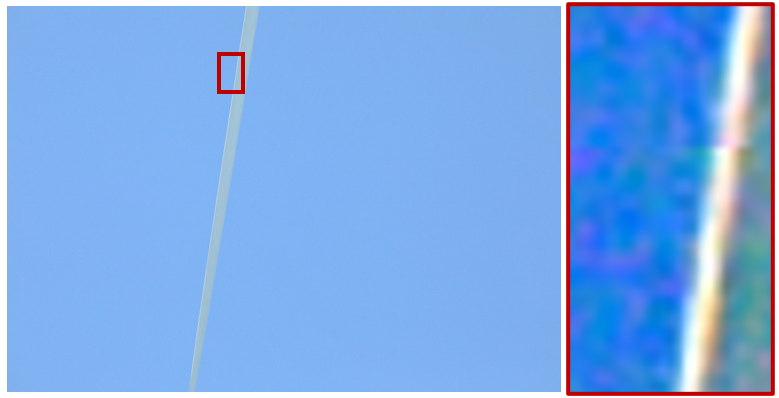}
\caption*{\centering \parbox{\linewidth}{\centering\fontsize{30}{36}\selectfont HP~\cite{hyperprior} \\ 34.71 / 0.9710 / 1.5334 }}
\end{subfigure}
&
\begin{subfigure}[b]{0.75\textwidth}
\includegraphics[width=\linewidth]{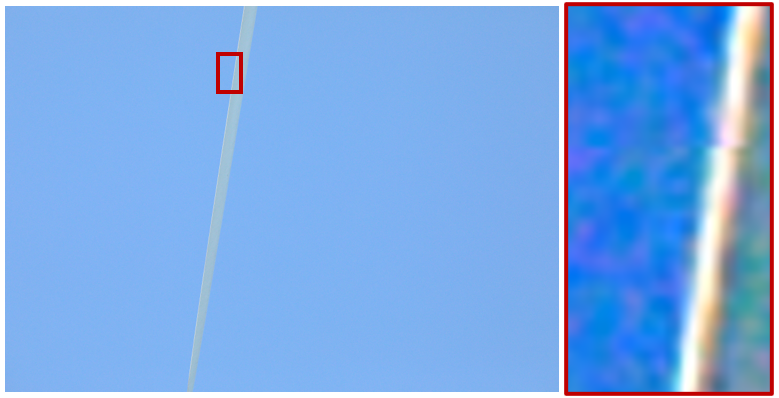}
\caption*{\centering \parbox{\linewidth}{\centering\fontsize{30}{36}\selectfont $L=3$ \\ 34.78 / 0.9716 / 1.5367 }}
\end{subfigure}
&
\begin{subfigure}[b]{0.75\textwidth}
\includegraphics[width=\linewidth]{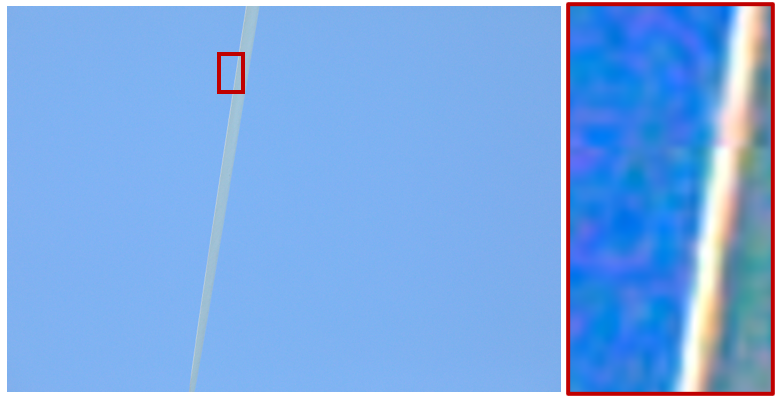}
\caption*{\centering \parbox{\linewidth}{\centering\fontsize{30}{36}\selectfont $L=4$ \\ 34.90 / 0.9727 / 1.5525 }}
\end{subfigure}
&
\begin{subfigure}[b]{0.75\textwidth}
\includegraphics[width=\linewidth]{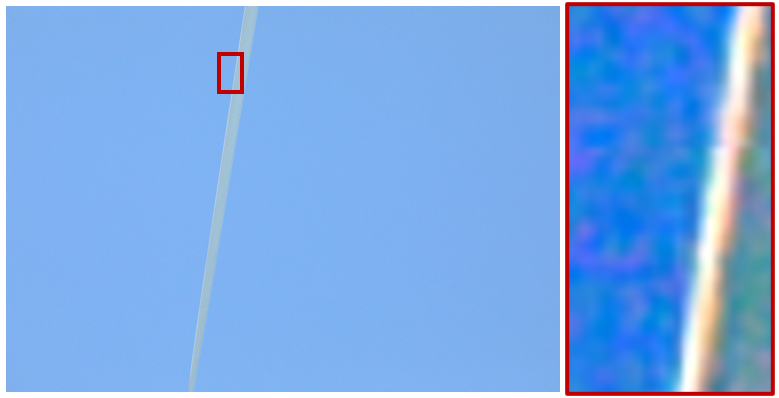}
\caption*{\centering \parbox{\linewidth}{\centering\fontsize{30}{36}\selectfont $L=5$ \\ 34.89 / 0.9725 / 1.5576 }}
\end{subfigure}

\\ & & & & \\

\begin{subfigure}[b]{0.75\textwidth}
\includegraphics[width=\linewidth]{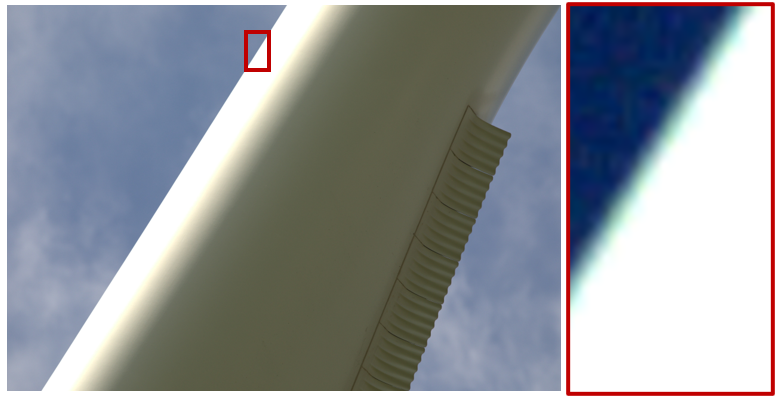}
\caption*{\centering \parbox{\linewidth}{\centering\fontsize{30}{36}\selectfont Original image \\ \textcolor{white}{.} }}
\end{subfigure}
&
\begin{subfigure}[b]{0.75\textwidth}
\includegraphics[width=\linewidth]{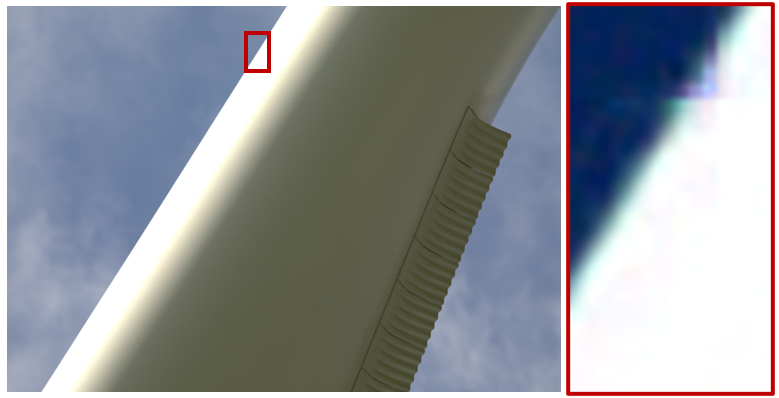}
\caption*{\centering \parbox{\linewidth}{\centering\fontsize{30}{36}\selectfont HP~\cite{hyperprior} \\ 43.05 / 0.9916 / 0.8910 }}
\end{subfigure}
&
\begin{subfigure}[b]{0.75\textwidth}
\includegraphics[width=\linewidth]{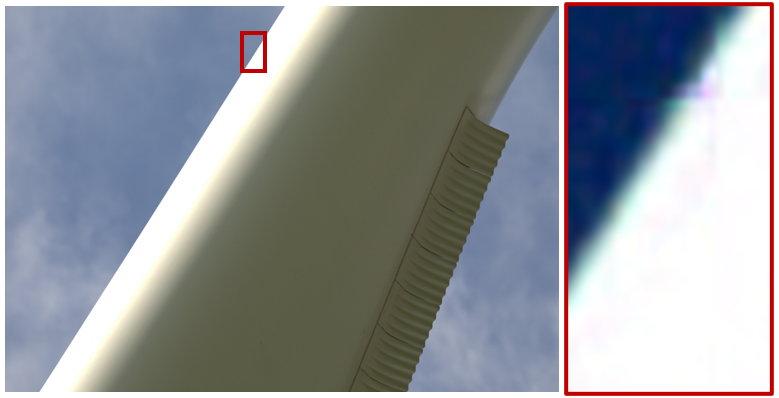}
\caption*{\centering \parbox{\linewidth}{\centering\fontsize{30}{36}\selectfont $L=3$ \\ 43.07 / 0.9918 / 0.8967 }}
\end{subfigure}
&
\begin{subfigure}[b]{0.75\textwidth}
\includegraphics[width=\linewidth]{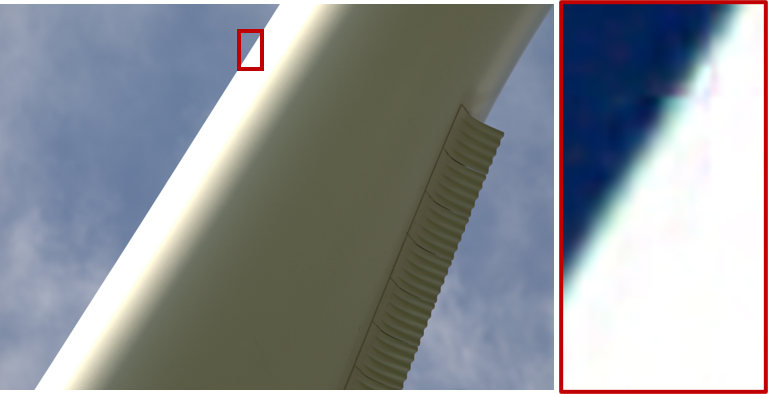}
\caption*{\centering \parbox{\linewidth}{\centering\fontsize{30}{36}\selectfont $L=4$ \\ 43.30 / 0.9920/ 0.9097 }}
\end{subfigure}
&
\begin{subfigure}[b]{0.75\textwidth}
\includegraphics[width=\linewidth]{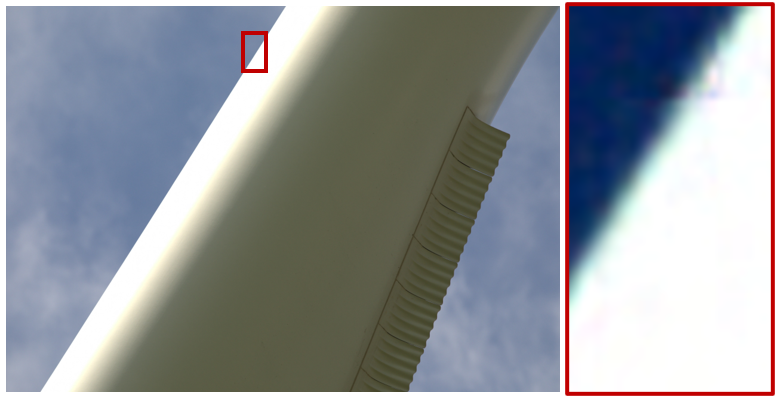}
\caption*{\centering \parbox{\linewidth}{\centering\fontsize{30}{36}\selectfont $L=5$ \\ 43.32 / 0.9920 / 0.9069 }}
\end{subfigure}

\\ & & & & \\

\begin{subfigure}[b]{0.75\textwidth}
\includegraphics[width=\linewidth]{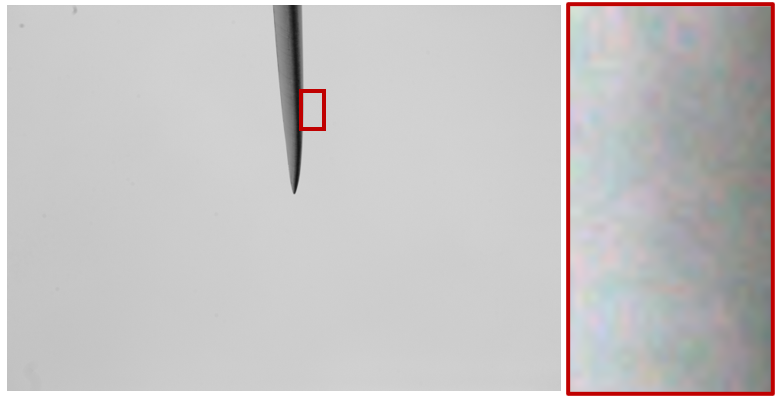}
\caption*{\centering \parbox{\linewidth}{\centering\fontsize{30}{36}\selectfont Original image \\ \textcolor{white}{.} }}
\end{subfigure}
&
\begin{subfigure}[b]{0.75\textwidth}
\includegraphics[width=\linewidth]{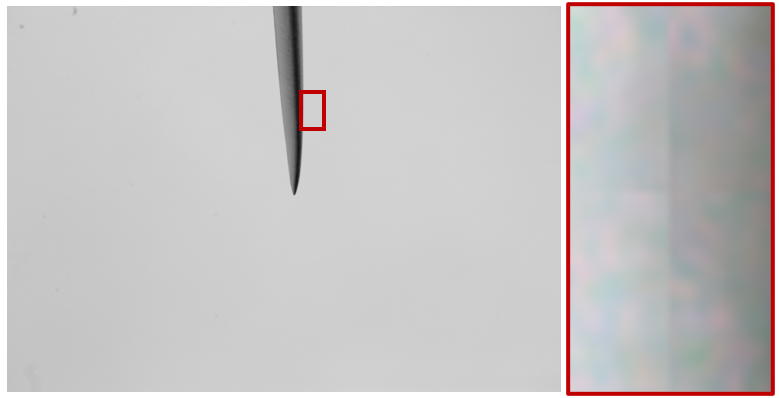}
\caption*{\centering \parbox{\linewidth}{\centering\fontsize{30}{36}\selectfont HP~\cite{hyperprior} \\ 40.91 / 0.9876 / 1.1691 }}
\end{subfigure}
&
\begin{subfigure}[b]{0.75\textwidth}
\includegraphics[width=\linewidth]{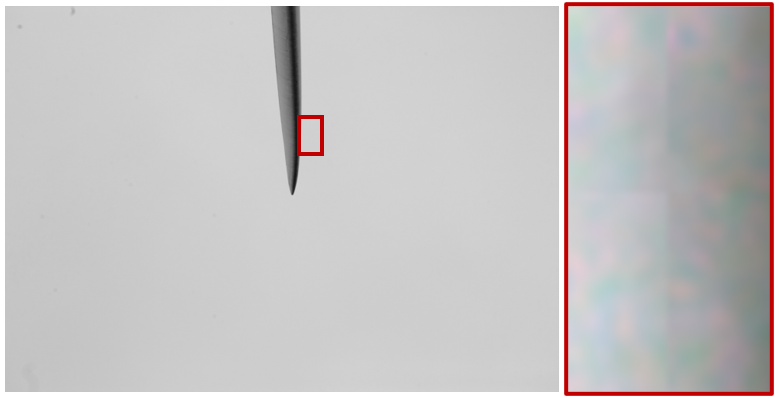}
\caption*{\centering \parbox{\linewidth}{\centering\fontsize{30}{36}\selectfont $L=3$ \\ 41.19 / 0.9886 / 1.2185 }}
\end{subfigure}
&
\begin{subfigure}[b]{0.75\textwidth}
\includegraphics[width=\linewidth]{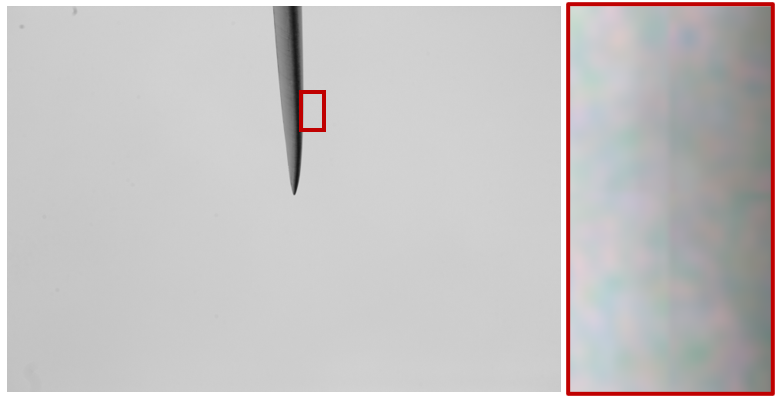}
\caption*{\centering \parbox{\linewidth}{\centering\fontsize{30}{36}\selectfont $L=4$ \\ 41.27 / 0.9889/ 1.2186 }}
\end{subfigure}
&
\begin{subfigure}[b]{0.75\textwidth}
\includegraphics[width=\linewidth]{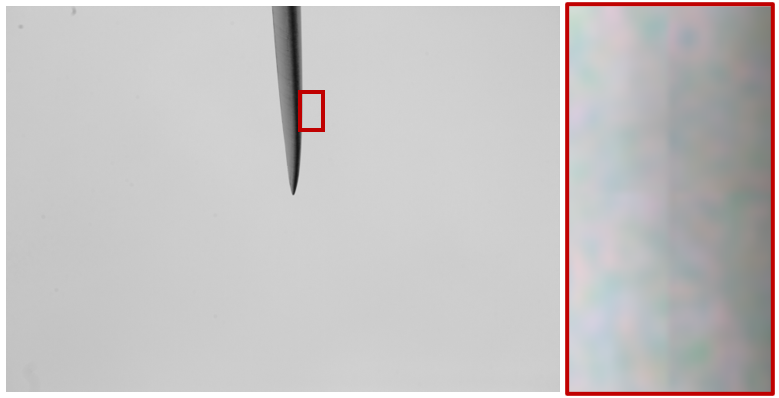}
\caption*{\centering \parbox{\linewidth}{\centering\fontsize{30}{36}\selectfont $L=5$ \\ 41.21 / 0.9887 / 1.2162 }}
\end{subfigure}

\end{tabular}}
\end{center} 
\vspace{-0.5cm}
\caption{\textbf{Visual comparison of distinct blade instances with respect to $L$}. Subcaptions denote the PSNR, MS-SSIM and bit/px. HP~\cite{hyperprior} employs $L=2$. Zoomed region contrast is increased in the images on the first and fourth rows.} \vspace{-0.4cm}
\label{fig:visual} 
\end{figure*}

\vspace{-0.35cm}
\subsection{Implementation Details} \label{sec:training}
\vspace{-0.1cm}

The model processes 256$\times$256 pixel patches using an NVIDIA GeForce RTX 3080 Ti. Training employs the Adam optimizer~\cite{adam} with an initial learning rate of $10^{-4}$. Two distortion metrics are used: Mean Squared Error (MSE) and negative Multi-Scale Structural SIMilarity (MS-SSIM)~\cite{msssim}. To ensure stable training and control the gradients, the rate terms have a minimum bound of $10^{-9}$. Overfitting prevention involves early-stopping~\cite{early} and diversified training data via random cropping from full-resolution images in each epoch.

\begin{figure*}[t!]
  \begin{subfigure}{0.5\textwidth}
    \begin{minipage}{\linewidth}
      \includegraphics[width=\linewidth]{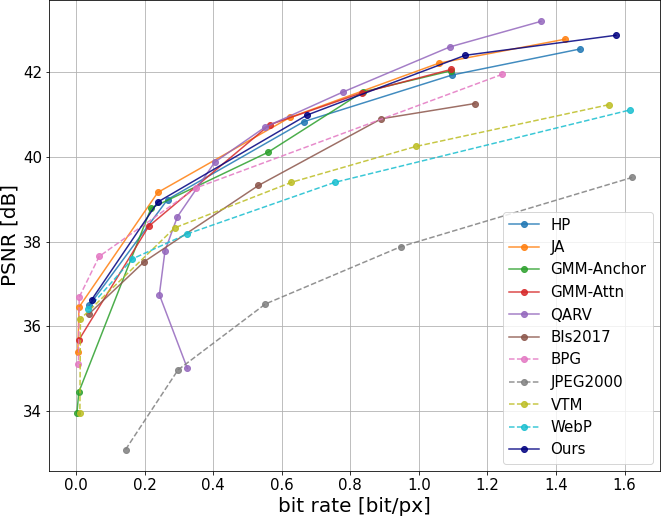}
      \vspace{-0.45cm}
      \subcaption{PSNR quality measure}
      \label{subfig:sota-psnr}
    \end{minipage}
  \end{subfigure}%
  \begin{subfigure}{0.5\textwidth}
    \begin{minipage}{\linewidth}
      \includegraphics[width=\linewidth]{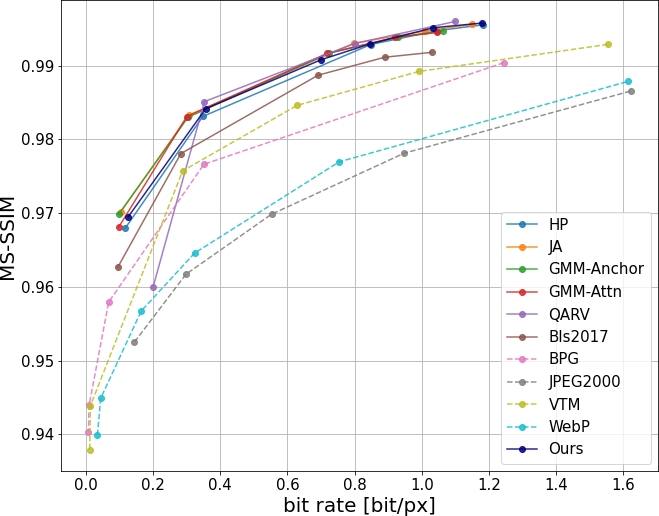}
      \vspace{-0.45cm}
      \subcaption{MS-SSIM quality measure}
      \label{subfig:sota-msssim}
    \end{minipage}
  \end{subfigure}   
    \vspace{-0.35cm}
  \caption{ \textbf{Test rate-distortion curves for distinct lossy coders}. Plots are divided by the distortion metric. Best viewed in color.} \label{fig:sota} \vspace{-0.4cm}
\end{figure*} 

\vspace{-0.25cm}
\subsection{Ablation Study} \label{sec:ablation}
\vspace{-0.1cm}

We conduct an ablation study of the lossy neural architecture trained on MSE for $L=\{3,4,5\}$; presented in Section~\ref{sec:model}. In particular, Fig.~\ref{fig:ablation} explores the performance of four distinct architectures on the validation set in terms of rate-distortion curves: the baseline extends HP~\cite{hyperprior} for any $L$, an architecture featuring equal dimensionality $M=150$ across all latent variables, an architecture employing a standard logistic prior, and a hybrid architecture integrating the two previous ones. 


Both architectural modifications significantly enhance compression performance, especially in scenarios with a large value of $L$, higher bit rates, or even more when both factors are combined. For $L=3$, this hybrid architecture demonstrates that an equal compression performance compared to the baseline is achieved, thus, the prior entropy network~\cite{hyperprior} is no longer required to achieve top-performing results. When $L=4$, we observe that for the highest bit rate model, the hybrid architecture surpasses the rest and, for $L=5$, this outperforming behaviour extends to all the rate-distortion curve, showing that the prior entropy network is indeed detrimental.

The hybrid architectures of $L=\{3,4,5\}$ and HP~\cite{hyperprior}, which corresponds to the baseline with $L=2$, are compared in Fig.~\ref{fig:model_l}. The evaluation is made depending on the distortion metric used during the training: Peak Signal-to-Noise Ratio (PSNR) for $d=$MSE and MS-SSIM for $d=1-$MS-SSIM.

Figure~\ref{subfig:model_l-psnr} illustrates that higher bit rates demand a larger $L$ for optimal results. Specifically, $L=\{4,5\}$ excel on highest bit rate scenarios, $L=3$ outperforms on intermediate bit rate cases, and comparable compression performance is exhibited for any $L$ at the lowest bit rate. This observed behavior can be attributed to the network's inherent struggle to minimize distortion versus reducing the bit rate. To elucidate, when increasing $\lambda$ and intensifying the requirement for image quality, the network necessitates more epochs to converge. Therefore, a higher image quality demand entails a more complex network, which translates in a larger $L$. 

A similar trend is noticeable in Fig.~\ref{subfig:model_l-msssim}. Nevertheless, the distinctions between the architectures are less pronounced in MS-SSIM, and its graphical representation is somewhat impaired. Within this figure, we offer a more comprehensive visualization by directly comparing the losses of each $L$, revealing that the optimal $L$ increases when the bit rate does. 

\vspace{-0.35cm}
\subsection{Visual Analysis} \label{sec:visual}
\vspace{-0.1cm}

To shed light on how generalized nested models impact the visual perspective, Fig.~\ref{fig:visual} depicts blade picture instances compressed with a distinct number of latent variables $L=\{2,3,4,5\}$. For $L=\{3,4,5\}$, the network incorporates the hybrid modifications discussed in Section~\ref{sec:ablation}. Because these pictures are high-resolution images, the figure focuses on the comparison of the distinct models on image patches with high frequency information, so we can facilitate its comparison. The contrast may be increased for the same reason.

To start with the first row, we can clearly see on the original image a perfect line boundary between the blade region and the blue sky. However, when the image is compressed using HP~\cite{hyperprior} ($L=2$), much noisy distortion is introduced on the blade; and also on the background. When applying generalized networks, this distortion is iteratively reduced when we increase $L$. Within the same bit rate range, by utilizing $L=3$, the distortion metrics are reduced and its perceptual view is enhanced. What is more, our 4-layer latent model substantially drops this distortion noise and even improves the quality measures. Note that for $L=5$, we reach an equal level of compression performance as $L=4$, showing the optimal layer depth for this rate-distortion trade-off is $L=4$.

Another common distortion iteratively reduced when incorporating additional latent variables is blocking artifacts; caused due to compressing the images through individual image patches. As seen from the second to the last row, nested models can selectively enhance those regions. In all cases, we obtain again $L=4$ as the optimal number of latent variables.

\begin{table*}[ht!]
\centering
\resizebox{17.8 cm}{!} {
\begin{tabular}{lcccccccccc}
\toprule
    & & \multicolumn{4}{c}{\textbf{Compression Time}} & & \multicolumn{4}{c}{\textbf{Decompression Time}} \\
    \multicolumn{1}{c}{Coding} & & \multicolumn{1}{c}{Minimum}  & \multicolumn{1}{c}{Maximum} & \multicolumn{1}{c}{Mean} & Slope & & \multicolumn{1}{c}{Minimum}  & \multicolumn{1}{c}{Maximum} & \multicolumn{1}{c}{Mean} & Slope  \\
    \multicolumn{1}{c}{scheme} & & time [s] & time [s] & time [s]  & growth [s/bit] & & time [s] & time [s] & time [s]  & growth [s/bit] \\
    \midrule  
   HP \cite{hyperprior} & &  5.76 &    6.17 &    5.96 &    0.25 &  &  6.51 &  15.60 &  12.83 &   5.57\\
           JA \cite{ja} & & 200.68 &  204.19 &  203.01 &    1.96 &  & 442.74 & 458.26 & 451.08 &   8.69 \\
   GMM-Anchor \cite{gmm} & & 227.91 &  235.96 &  233.04 &    5.51 & & 480.62 & 495.27 & 487.40 &  10.02 \\
     GMM-Attn \cite{gmm} & & 247.97 &  258.12 &  252.81 &    6.78 & & 500.60 & 523.14 & 508.58 &  15.06 \\
      Bls2017 \cite{bls2017} & &  0.27 &    0.35 &    0.30 &    0.07  & &   2.85 &  11.94 &   9.12 &   7.36 \\
      QARV \cite{qarv} &  & 371.08 & 374.68 & 373.28 &   2.21   &  & 238.08 & 240.99 & 239.70 &   1.79   \\
          BPG \cite{bpg} & &  2.47 &    8.80 &    4.75 &    3.25 &  & 1.08 &  12.78 &   5.91 &   6.01 \\
     JPEG2000 \cite{J2K} & &  7.02 &    7.03 &    7.02 &    0.01 &  & 24.40 &  58.48 &  38.92 &  16.66 \\
          VTM \cite{vtm} & &102.57 & 9943.02 & 4584.58 & 5309.94 &  & 1.42 &   4.32 &   2.73 &   1.57 \\
         WebP \cite{webp} &  & 2.49 &    9.40 &    4.17 &    3.59 &  & 1.26 &  10.87 &   5.56 &   4.99 \\
         Ours  & &  9.10 & 13.12 & 10.73 &   2.71  &  &  9.51 & 18.82 & 15.24 &   6.26  \\ 
\bottomrule
\end{tabular}
}
\vspace{-0.3cm} 
\caption{\textbf{Running time comparison of a full-resolution image} for distinct state-of-the-art lossy coders.} \label{tab:time} \vspace{-0.55cm}
\end{table*}

\vspace{-0.35cm}
\subsection{Quantitative Comparison} \label{sec:quantitative}
\vspace{-0.1cm}

In order to contextualize our approach, we present a concise comparison with state-of-the-art lossy coders in Fig.~\ref{fig:sota}. To this end, we selected the optimal layer depth $L$ over the validation set (Fig.~\ref{fig:model_l}) for each rate-distortion model to build our proposed lossy coder; which is now evaluated on the test set. 

Learning-based coders highlight superior performance than traditional ones, except for BPG~\cite{bpg} evaluated on PSNR, which surpasses the rest on low bit rate scenarios. Our approach excels on compression performance in both quality measures, accomplishing comparable results to top-performing autoregressive coders, such as JA~\cite{ja}. However, JA's autoregressive nature weakens its applicability due to its high computational cost, as we will see in Section~\ref{sec:time}. Despite outperforming our model in high bit rates, QARV's~\cite{qarv} embedding layer is not sufficient to generalize for distinct rate-distortion terms, therefore, it performs poorly on low bit rates. In addition to that, its computational cost is higher than a typical autoregressive model, as discussed later.

Focusing on the baseline codecs of our proposed generalized networks, we substantially outperform in compression performance Bls2017~\cite{bls2017}, which corresponds to $L=1$, and show significant improvement also compared to HP~\cite{hyperprior} (which is $L=2$). As mentioned in Section~\ref{sec:ablation}, generalized nested models significantly enhance the compression performance on high bit rate scenarios, when the image quality requirements imply a large $L$ as optimal. In this range of image quality is relevant to accomplish a high compression performance, because they ensure an acceptable image quality to perform defect detection during blade assessments.

\vspace{-0.25cm}
\subsection{Computational Cost} \label{sec:time}
\vspace{-0.1cm}

Table~\ref{tab:time} shows the computational cost of coding a wind turbine blade image using an NVIDIA RTX 3080 Ti GPU and a 20-core Intel Core i9-10900 KF processor at 3.70 GHz, assuming pre-loaded model weights to simulate a drone inspection.

We analyzed the computational cost of adding an extra nested latent variable. Using HP~\cite{hyperprior} as a reference, our generalized hyperprior model shows a linear increase in cost with each new latent variable, as its parameter count grows linearly and the architecture structure is a pure extension. Notably, a new latent variable results in an additional demand of approximately $4$s during the encoding and $2$s during decoding. 


In Table~\ref{tab:time}, we showcase the minimum, maximum, and average coding run times in seconds. Recognizing the typical trend of increased computational time with higher bit rates attributed to the entropy coder, we also present the growth slope observed as the bit rate increases. Our objective is to achieve a high-performing coder characterized by swift compression and decompression times, exhibiting minimal sensitivity to the desired bit rate; signified by a low slope growth.

Our nested latent variable model meets these criteria effectively, demonstrating outstanding compression performance, while not significantly increasing the coding time, except compared to Bls2017~\cite{bls2017}. It exhibits modest compression slope growth and maintains reasonably low average decompression time. In contrast, autoregressive models prove to be considerably slower due to their sequential contextual prediction process. Notably, QARV~\cite{qarv}, despite not being autoregressive, fails to strike a favorable balance between coding performance and computational cost. Consequently, our approach matches autoregressive models in compression performance while reducing computational cost.




%% file: Chapters/Conclusion.tex
\vspace{-0.35cm}
\section{Conclusion}
\vspace{-0.25cm}

This paper extends the scope of the deep hyperprior for neural lossy coding, introducing a versatile $L$-level nested latent model. Our method captures the intricate dependencies among latent variables with greater fidelity and marked compression improvement. By carefully designing the architecture and selecting the optimal layer depth depending on the rate-distortion trade-off, these generalized models surpass the hyperprior performance without a trainable prior and successfully approximate autoregressive models, accomplishing state-of-the-art results while reducing substantially the computational cost. Our framework effectiveness is solidified through empirical evaluation in a real-world context. Specifically, we have demonstrated its applicability within visual wind turbine inspection data by yielding compelling results, serving as a testament to its robustness and practicality.


%% file: main.bbl
\begin{thebibliography}{10}

\bibitem{J2K}
A.~Skodras, C.~Christopoulos, and T.~Ebrahimi,
\newblock ``The {JPEG} 2000 still image compression standard,''
\newblock {\em SPM}, vol. 18, no. 5, pp. 36--58, 2001.

\bibitem{webp}
{Google},
\newblock ``{WebP: Compression techniques},'' \url{https://developers.google.com/speed/webp/docs/compression}, 2017.

\bibitem{hevc}
G.J. Sullivan, J.R. Ohm, W.J. Han, and T.~Wiegand,
\newblock ``Overview of the high efficiency video coding (hevc) standard,''
\newblock {\em TCSVT}, vol. 22, no. 12, pp. 1649--1668, 2012.

\bibitem{bpg}
F.~Bellard,
\newblock ``{BPG} image format,''
\newblock {\em SPIC}, 2015.

\bibitem{vtm}
{Joint Video Exploration Team (JVET)},
\newblock ``{VVC VTM},'' Software available at \url{https://vcgit.hhi.fraunhofer.de/jvet/VVCSoftware_VTM}.

\bibitem{review}
S.~Ma, X.~Zhang, C.~Jia, Z.~Zhao, S.~Wang, and S.~Wang,
\newblock ``Image and video compression with neural networks: A review,''
\newblock {\em TCSVT}, vol. 30, no. 6, pp. 1683--1698, 2019.

\bibitem{review2}
N.~Anantrasirichai and D.~Bull,
\newblock ``Artificial intelligence in the creative industries: a review,''
\newblock {\em AIR}, 2022.

\bibitem{rnn}
G.~Toderici, S.~M. O'Malley, S.~J. Hwang, D.~Vincent, D.~Minnen, S.~Baluja, M.~Covell, and R.~Sukthankar,
\newblock ``Variable rate image compression with recurrent neural networks,''
\newblock {\em ICLR}, 2016.

\bibitem{review3}
Y.~Yang, S.~Mandt, L.~Theis, et~al.,
\newblock ``An introduction to neural data compression,''
\newblock {\em FTCGV}, vol. 15, no. 2, pp. 113--200, 2023.

\bibitem{bls2017}
J.~Ball{\'e}, V.~Laparra, and E.~P. Simoncelli,
\newblock ``End-to-end optimized image compression,''
\newblock {\em ICLR}, 2017.

\bibitem{hyperprior}
J.~Ball{\'e}, D.~Minnen, S.~Singh, S.~J. Hwang, and N.~Johnston,
\newblock ``Variational image compression with a scale hyperprior,''
\newblock {\em ICLR}, 2018.

\bibitem{l3}
Y.~Hu, W.~Yang, and J.~Liu,
\newblock ``Coarse-to-fine hyper-prior modeling for learned image compression,''
\newblock in {\em AAAI}, 2020.

\bibitem{wacv}
Z.~Duan, M.~Lu, Z.~Ma, and F.~Zhu,
\newblock ``Lossy image compression with quantized hierarchical vaes,''
\newblock in {\em WACV}, 2023.

\bibitem{qarv}
Z.~Duan, M.~Lu, J.~Ma, Y.~Huang, Z.~Ma, and F.~Zhu,
\newblock ``{QARV}: Quantization-aware resnet {VAE} for lossy image compression,''
\newblock {\em TPAMI}, vol. 46, no. 1, pp. 436--450, 2024.

\bibitem{ja}
D.~Minnen, J.~Ball{\'e}, and G.~D. Toderici,
\newblock ``Joint autoregressive and hierarchical priors for learned image compression,''
\newblock {\em NeurIPS}, 2018.

\bibitem{context}
J.~Lee, S.~Cho, and S.~K. Beack,
\newblock ``Context-adaptive entropy model for end-to-end optimized image compression,''
\newblock {\em ICLR}, 2019.

\bibitem{context2}
D.~He, Y.~Zheng, B.~Sun, Y.~Wang, and H.~Qin,
\newblock ``Checkerboard context model for efficient learned image compression,''
\newblock in {\em CVPR}, 2021.

\bibitem{context_icassp}
A.~Meyer and A.~Kaup,
\newblock ``A novel cross-component context model for end-to-end wavelet image coding,''
\newblock in {\em ICASSP}, 2023.

\bibitem{mlic}
W.~Jiang, J.~Yang, Y.~Zhai, P.~Ning, F.~Gao, and R.~Wang,
\newblock ``Mlic: Multi-reference entropy model for learned image compression,''
\newblock in {\em ACM Multimedia}, 2023.

\bibitem{easn}
C.~Shin, H.~Lee, H.~Son, S.~Lee, D.~Lee, and S.~Lee,
\newblock ``Expanded adaptive scaling normalization for end to end image compression,''
\newblock in {\em ECCV}, 2022.

\bibitem{attention}
H.~Liu, T.~Chen, P.~Guo, Q.~Shen, X.~Cao, Y.~Wang, and Z.~Ma,
\newblock ``Non-local attention optimized deep image compression,''
\newblock {\em TIP}, vol. 30, no. 2, pp. 3179--3191, 2021.

\bibitem{gmm}
Z.~Cheng, H.~Sun, M.~Takeuchi, and J.~Katto,
\newblock ``Learned image compression with discretized gaussian mixture likelihoods and attention modules,''
\newblock in {\em CVPR}, 2020.

\bibitem{jiro}
J.~Liu, H.~Sun, and J.~Katto,
\newblock ``Learned image compression with mixed transformer-cnn architectures,''
\newblock in {\em CVPR}, 2023.

\bibitem{deep-latent1}
L.~Maal{\o}e, M.~Fraccaro, V.~Li{\'e}vin, and O.~Winther,
\newblock ``{BIVA}: A very deep hierarchy of latent variables for generative modeling,''
\newblock {\em NeurIPS}, 2019.

\bibitem{deep-latent2}
C.~K. S{\o}nderby, T.~Raiko, L.~Maal{\o}e, S.~K. S{\o}nderby, and O.~Winther,
\newblock ``Ladder variational autoencoders,''
\newblock {\em NeurIPS}, 2016.

\bibitem{time}
F.~Pakdaman and M.~Gabbouj,
\newblock ``Comprehensive complexity assessment of emerging learned image compression on {CPU} and {GPU},''
\newblock in {\em ICASSP}, 2023.

\bibitem{PerezGonzaloIcip2023}
R.~P\'erez-Gonzalo, A.~Espersen, and A.~Agudo,
\newblock ``Robust wind turbine blade segmentation from {RGB} images in the wild,''
\newblock in {\em ICIP}, 2023.

\bibitem{ai-drone}
A.~Shihavuddin, X.~Chen, V.~Fedorov, A.~Nymark~Christensen, N.~Andre Brogaard~Riis, K.~Branner, A.~Bjorholm~Dahl, and R.~Reinhold~Paulsen,
\newblock ``Wind turbine surface damage detection by deep learning aided drone inspection analysis,''
\newblock {\em Energies}, vol. 12, no. 4, pp. 676, 2019.

\bibitem{downtime}
S.~Nandipati, A.~N. Nichenametla, and A.~L. Waghmare,
\newblock ``Cost-effective maintenance plan for multiple defect types in wind turbine blades,''
\newblock in {\em RMS}, 2018.

\bibitem{downtime2}
A.~Gonz{\'a}lez-Gonz{\'a}lez, A.~J. Cortadi, D.~Galar, and L.~Ciani,
\newblock ``Condition monitoring of wind turbine pitch controller: A maintenance approach,''
\newblock {\em Measurement}, vol. 123, pp. 80--93, 2018.

\bibitem{ans}
J.~Duda, K.~Tahboub, N.~J. Gadgil, and E.~J. Delp,
\newblock ``The use of asymmetric numeral systems as an accurate replacement for huffman coding,''
\newblock in {\em PCS}, 2015.

\bibitem{pixelcnn}
A.~Oord, N.~Kalchbrenner, O.~Vinyals, L.~Espeholt, A.~Graves, and K.~Kavukcuoglu,
\newblock ``Conditional image generation with pixelcnn decoders,''
\newblock in {\em NeurIPS}, 2016.

\bibitem{channel}
D.~Minnen and S.~Singh,
\newblock ``Channel-wise autoregressive entropy models for learned image compression,''
\newblock in {\em ICIP}, 2020.

\bibitem{checkerboard}
D.~He, Y.~Zheng, B.~Sun, Y.~Wang, and H.~Qin,
\newblock ``Checkerboard context model for efficient learned image compression,''
\newblock in {\em CVPR}, 2021.

\bibitem{approximate}
R.~Child,
\newblock ``Very deep vaes generalize autoregressive models and can outperform them on images,''
\newblock in {\em ICLR}, 2021.

\bibitem{adam}
D.~P. Kingma and J.~Ba,
\newblock ``Adam: A method for stochastic optimization,''
\newblock {\em ICLR}, 2015.

\bibitem{msssim}
Z.~Wang, E.~P. Simoncelli, and A.~C. Bovik,
\newblock ``Multiscale structural similarity for image quality assessment,''
\newblock in {\em ACSSC}, 2003.

\bibitem{early}
L.~Prechelt,
\newblock ``Early stopping-but when?,''
\newblock in {\em Neural Networks: Tricks of the trade}. 1998.

\end{thebibliography}
